\documentclass{article}
\usepackage{spconf,amsmath,graphicx}
\usepackage{booktabs,multirow,array}
\usepackage[table]{xcolor} 
\usepackage{tabularx}
\usepackage{rotating}
\usepackage{changepage}
\usepackage{subcaption}
\usepackage{blindtext}
\pdfoutput=1

\title{Learning a Deep Convolution Network with Turing Test Adversaries for Microscopy Image Super Resolution}

\name{Francis~Tom$^\star$, Himanshu~Sharma$^\dagger$, Dheeraj~Mundhra$^\dagger$, Tathagato~Rai~Dastidar$^\dagger$, Debdoot~Sheet$^\star$}
\address{$^\star$Indian Institute of Technology Kharagpur, Kharagpur, WB 721302, India \\
    \texttt{debdoot@ee.iitkgp.ac.in}\\
$^\dagger$SigTuple Technologies Private Limited, Bengaluru, KA 560102, India\\
  \texttt{trd@sigtuple.com}}

\begin{document}
\maketitle

\begin{abstract}
Adversarially trained deep neural networks have significantly improved performance of single image super resolution, by hallucinating photorealistic local textures, thereby greatly reducing the perception difference between a real high resolution image and its super resolved (SR) counterpart. However, application to medical imaging requires preservation of diagnostically relevant features while refraining from introducing any diagnostically confusing artifacts. We propose using a deep convolutional super resolution network (SRNet) trained for (i) minimising reconstruction loss between the real and SR images, and (ii) maximally confusing learned relativistic visual Turing test (rVTT) networks to discriminate between (a) pair of real and SR images (T1) and (b) pair of patches in real and SR selected from region of interest (T2). The adversarial loss of T1 and T2 while backpropagated through SRNet helps it learn to reconstruct pathorealism in the regions of interest such as white blood cells (WBC) in peripheral blood smears or epithelial cells in histopathology of cancerous biopsy tissues, which are experimentally demonstrated here. Experiments performed for measuring signal distortion loss using peak signal to noise ratio (pSNR) and structural similarity (SSIM) with variation of SR scale factors, impact of rVTT adversarial losses, and impact on reporting using SR on a commercially available artificial intelligence (AI) digital pathology system substantiate our claims.
\end{abstract}

\section{Introduction}

\begin{figure}[t]
\begin{center}
    \begin{subfigure}[b]{0.48\textwidth}
        \centering
        \includegraphics[width=\textwidth]{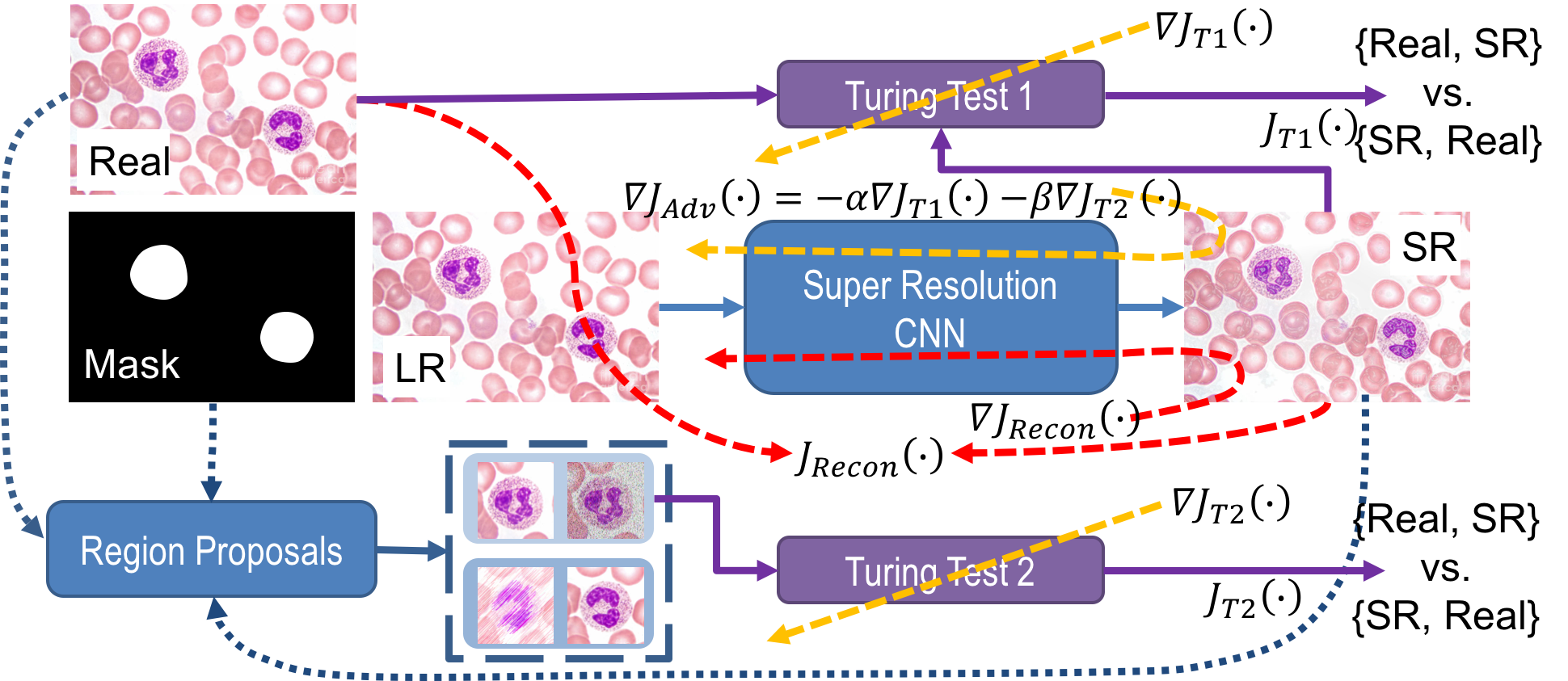}
        \caption{Overview of the adversarial learning process with Turing tests}
        \label{fig1a}
    \end{subfigure}
    \begin{subfigure}[b]{0.09\textwidth}
        \centering
        \includegraphics[trim=84 80 80 80, clip, width=\textwidth]{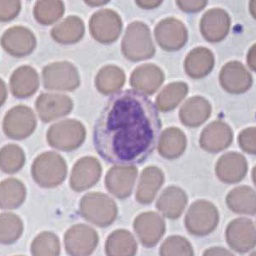}
        \caption{Real}
        \label{fig1b}
    \end{subfigure}
    \begin{subfigure}[b]{0.09\textwidth}
        \centering
        \includegraphics[trim=84 80 80 80, clip, width=\textwidth]{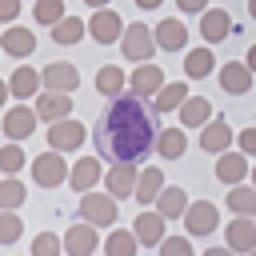}
        \caption{Bicubic}
        \label{fig1c}
    \end{subfigure}
    \begin{subfigure}[b]{0.09\textwidth}
        \centering
        \includegraphics[trim=84 80 80 80, clip, width=\textwidth]{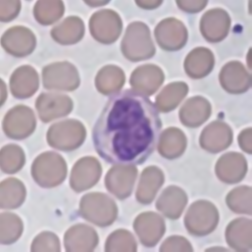}
        \caption{SRNet}
        \label{fig1d}
    \end{subfigure}
        \begin{subfigure}[b]{0.09\textwidth}
        \centering
        \includegraphics[trim=84 80 80 80, clip, width=\textwidth]{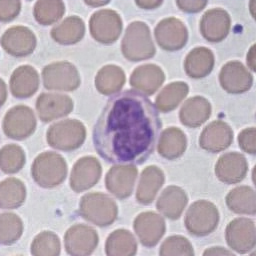}
        \caption{T1}
        \label{fig1e}
    \end{subfigure}
    \begin{subfigure}[b]{0.09\textwidth}
        \centering
        \includegraphics[trim=84 80 80 80, clip, width=\textwidth]{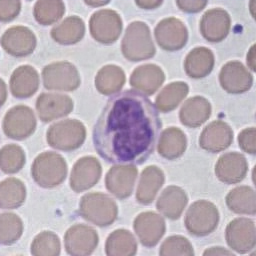}
        \caption{T1+T2}
        \label{fig1f}
    \end{subfigure}
    \begin{subfigure}[b]{0.09\textwidth}
        \centering
        \includegraphics[trim=24 120 150 80, clip, width=\textwidth]{Figures/Fig1/original.jpg}
        \caption{Real}
        \label{fig1g}
    \end{subfigure}
    \begin{subfigure}[b]{0.09\textwidth}
        \centering
        \includegraphics[trim=24 120 150 80, clip, width=\textwidth]{Figures/Fig1/bicubic.jpg}
        \caption{Bicubic}
        \label{fig1h}
    \end{subfigure}
    \begin{subfigure}[b]{0.09\textwidth}
        \centering
        \includegraphics[trim=24 120 150 80, clip, width=\textwidth]{Figures/Fig1/sr.png}
        \caption{SRNet}
        \label{fig1i}
    \end{subfigure}
        \begin{subfigure}[b]{0.09\textwidth}
        \centering
        \includegraphics[trim=24 120 150 80, clip, width=\textwidth]{Figures/Fig1/srt1.png}
        \caption{T1}
        \label{fig1j}
    \end{subfigure}
    \begin{subfigure}[b]{0.09\textwidth}
        \centering
        \includegraphics[trim=24 120 150 80, clip, width=\textwidth]{Figures/Fig1/srt1t2.png}
        \caption{T1+T2}
        \label{fig1k}
    \end{subfigure}
\caption{Learning a super-resolution CNN for microscopy using relativistic visual Turing tests (T1 and T2), the results obtained for super resolving by 16$\times$ and comparison with bicubic interpolation. Recovery of cytoplasmic texture and nuclear chromatin is evident in WBCs (b-f) while does not significantly impact relatively smooth textured RBCs (g-k).}
\label{fig:resultsnapshot}
\end{center}
\end{figure}

Single image super resolution (SISR) aims at estimating a high resolution (HR) image from a low resolution (LR) image. Image super resolution (SR) techniques can be used in microscopy to enhance the resolution of images acquired at a lower magnification thereby to reveal fine structures that could otherwise only be observed using a higher magnification lens. Accordingly SR images can be used to diagnose from images captured using a lower magnification, with diagnostic precision matching to that of images at a higher magnification, thereby facilitating to reduce the image acquisition time per slide significantly. SR being an ill-posed inverse problem gets challenging for high scaling factors, often affecting diagnostically relevant details such as texture in the SR images being absent. Fig.~\ref{fig:resultsnapshot} illustrates a simple example of failure to reproduce such fine details when learning a SR network (SRNet) using only distortion losses like the mean squared error (MSE), which fail to capture intricate details.

\begin{figure*}[!ht]
  \begin{center}
    \begin{subfigure}[b]{0.48\textwidth}
        \centering
        \includegraphics[width=\textwidth]{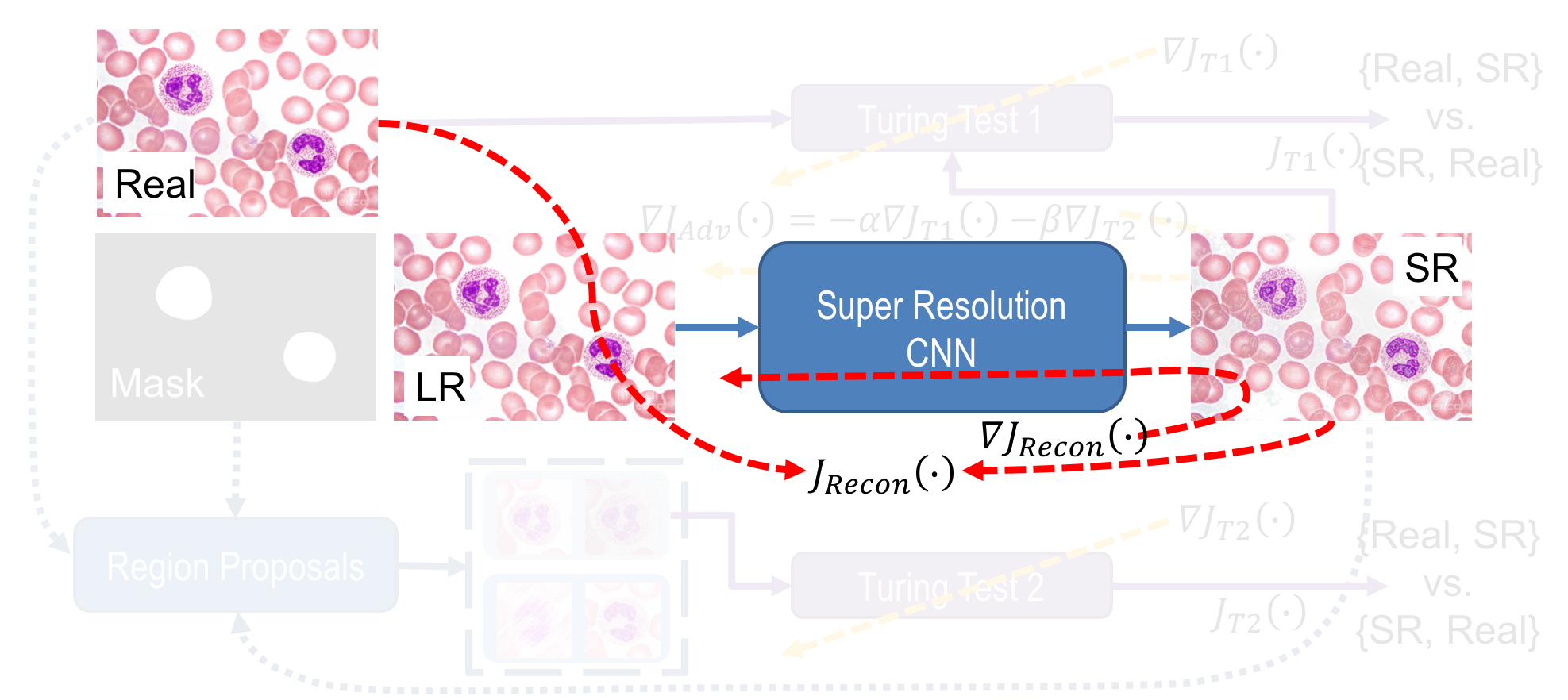}
        \caption{Stage 1: Training SRNet to minimize reconstruction error.}
        \label{fig2a}
    \end{subfigure}
    \begin{subfigure}[b]{0.48\textwidth}
        \centering
        \includegraphics[width=\textwidth]{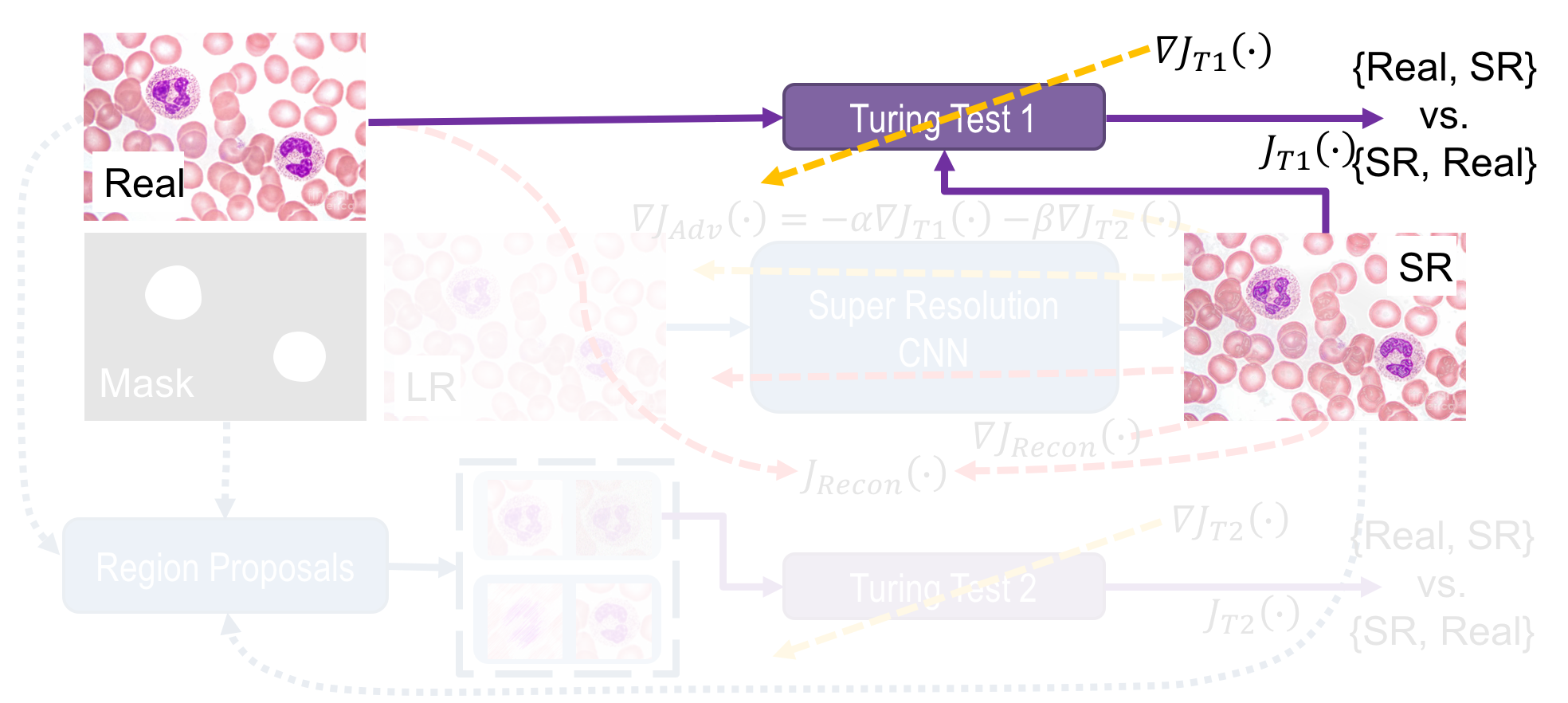}
        \caption{Stage 2: Training rVTT (T1) network on whole image.}
        \label{fig2b}
    \end{subfigure}
    \begin{subfigure}[b]{0.48\textwidth}
        \centering
        \includegraphics[width=\textwidth]{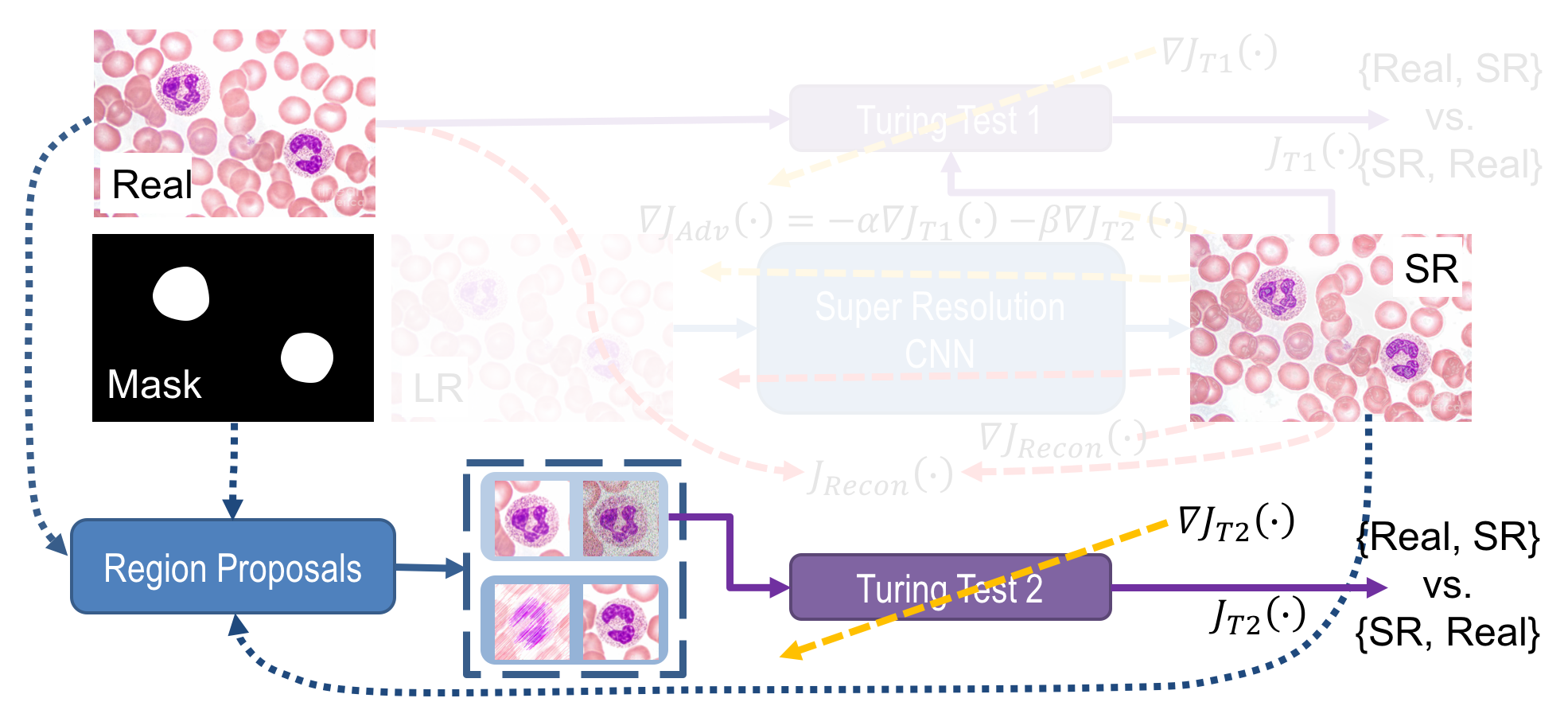}
        \caption{Stage 3: Training rVTT (T2) network on region of interest.}
        \label{fig2c}
    \end{subfigure}
    \begin{subfigure}[b]{0.48\textwidth}
        \centering
        \includegraphics[width=\textwidth]{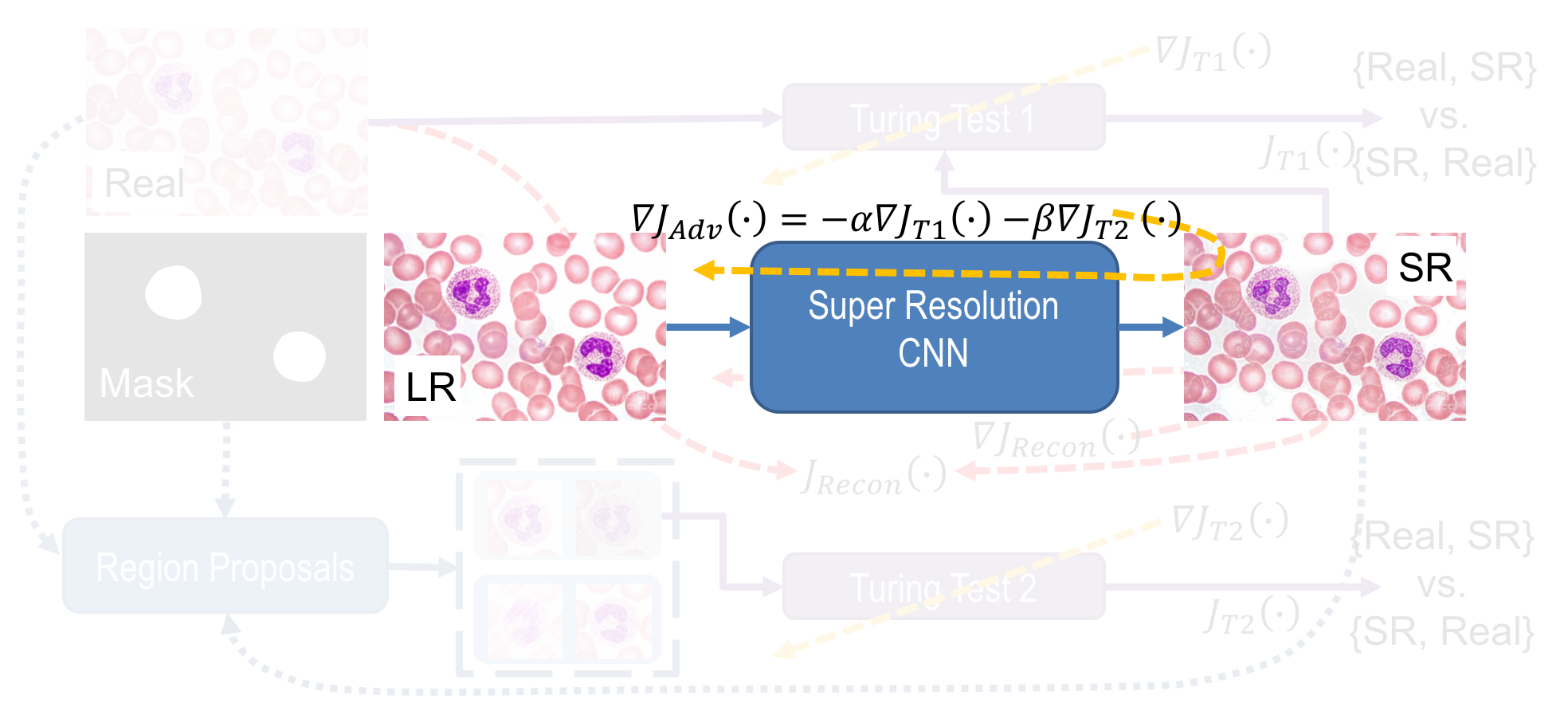}
        \caption{Stage 4: Training SRNet with adversarial loss.}
        \label{fig2d}
    \end{subfigure}
  \end{center}
\caption{Framework for microscopy image super resolution involving two relativistic visual Turing test (rVTT) networks.}
\label{fig:trainingprocess}
\end{figure*}

\section{Prior Art}

SISR has been traditionally solved using analytic or iterative methods that are physics-driven~\cite{srreview}. Recent works use deep learning as a data driven mean to enhance optical microscopy resolution~\cite{DL_microscopy} by employing mean square error with an edge weighting term as the loss function to be minimized. Another approach~\cite{DL_microscopy_mobile} employs a convolutional neural network (CNN) for the purpose but also suffers from the same limitation. Recent attempts via introduction of adversarial learning working in tandem with minimization of reconstruction loss have been able to recover fine texture details in natural images~\cite{SRGAN,ESRGAN}. However clinical grade microscopy being susceptible to region of interest (ROI) requires learning of specific grades of texture representation to be restored by the network within different regions in the image. Here we present its possibility.

\section{Method}
\label{sec:Method}

The goal is to train a super resolution neural network (SRNet) ($G(\cdot)$) that estimates for a given LR input image ( $\mathbf{I}_{LR}$) its corresponding SR counterpart ($\mathbf{I}_{SR}$) that closely resembles the real HR image ($\mathbf{I}_{HR}$). To achieve this, we propose a four stage learning process (\textbf{Stage 1}) train $G_{\theta_G}(\cdot)$ while optimizing its parameters ${\theta_G}$ with the objective to minimize reconstruction loss $J_{Recon}\left(\mathbf{I}_{SR}-\mathbf{I}_{HR}\right)$ utilizing $\nabla J_{Recon}(\cdot)$ to update $\theta_G$ as presented in Fig~\ref{fig2a}. Subsequently, (\textbf{Stage 2}) another CNN termed as the relativistic visual Turing test (rVTT) ($T1(\cdot)$) is trained with the objective to discriminate between the SR vs. Real HR image when presented with a matching pair of such images with shuffled order with the objective to be able to identify them correctly thereby minimizing $J_{T1}(\cdot)$ while updating parameters of T1 with $\nabla J_{T1}(\cdot)$ as illustrated in Fig.~\ref{fig2b}. This is used to quantify pair wise subtle difference in image perception which is a key factor different from distortion loss quantified in $J_{Recon}(\cdot)$~\cite{perception_distortion_tradeoff} and is inline with the philosophy in \cite{RAGAN,ESRGAN}. Next (\textbf{Stage 3}) we train another rVTT ($T2(\cdot)$) with the objective to be able to quantify the perception difference between SR vs. Real HR in diagnostically relevant ROI such as white blood cells (WBC) in peripheral blood smears or epithelial cells in histopathology of tissue biopsy colected from metaplastic regions. T2 is trained to minimizing $J_{T2}(\cdot)$ while updating its parameters with $\nabla J_{T2}(\cdot)$ as illustrated in Fig.~\ref{fig2c}. The region proposal finder relies on ROI masks provided as ground truth along with the images. Here we differ from~\cite{ESRGAN} since in pathological investigation microscopy the quantum of texture details varies with cells and tissue structure and in general the density of pathologically alarming cells being low~\cite{bancroft2008theory}, a single rVTT such as $T1(\cdot)$ alone is not able to properly encapsulate texture perception for such trace occurring cells. Finally (\textbf{Stage 4}) the objective being to update $G(\cdot)$ such that it can mimic in SR images the relativistic perception of global and ROI specific texture evident in Real HR images, we once again update $\theta_G$ with $\nabla J_{Adv}(\cdot)$ derived from the adversarial loss $J_{Adv}(\cdot)$ as presented in Fig.~\ref{fig2d}. On achieving this ability of $G(\cdot)$, it would essentially lead to maximization of $J_{T1}(\cdot)$ and $J_{T2}(\cdot)$ which forms the essence of adversarial learning.

\textbf{Architecture:} $G(\cdot)$ is similar to the one in \cite{ESRGAN} which features residual-in-residual dense blocks followed by strided convolutions for upsampling. $T1(\cdot)$ and $T2(\cdot)$ are also similar to the one in ~\cite{ESRGAN} and are modified versions of the VGG architecture~\cite{VGG}, with leaky ReLU non-linearity.

\textbf{Loss function:} The Stage 1 loss $J_{Recon}(\cdot)$ is defined as

\begin{equation}
    J_{Recon} = \eta |\mathbf{I}_{SR}-\mathbf{I}_{HR}|+L_{perception}(\cdot)
\end{equation}

\noindent where $L_{perception}(\cdot)$ is defined as the VGG perception loss detailed in~\cite{perceptual}. The Stage 2 and Stage 3 loss functions are similar to as proposed in ~\cite{RAGAN}.

\begin{multline}
 J_{T1} = -{E}_{{I}_{HR}}[\log(T1({I}_{HR}, {I}_{SR}))] \\- {E}_{{I}_{SR}}[\log(1 - T1({I}_{SR}, {I}_{HR}))]   
\end{multline}

\noindent where ${E}_{{I}_{HR}}(\cdot)$ denotes expectation over real HR images used in a mini-batch and ${E}_{{I}_{SR}}(\cdot)$ denotes expectation over SR images in a mini-batch.

\begin{multline}
 J_{T2} = -{E}_{{I}_{HR}}[\log(T1({x}_{HR}, {x}_{SR}))] \\- {E}_{{I}_{SR}}[\log(1 - T1({x}_{SR}, {x}_{HR}))].   
\end{multline}

\noindent where $\mathbf{x}_{HR}$ and $\mathbf{x}_{SR}$ are the image patches corresponding to ROI selected as in Fig.\ref{fig2c}. The adversarial loss $J_{Adv}(\cdot)$ in Stage 4 is defined in Fig.~\ref{fig2d}.

\subsubsection{Experiments, Results and Discussion}

\textbf{Dataset:} We evaluate the performance on three datasets: 

\textbf{ALL-IDB} \cite{ALL-IDB} where first 33 out of 108 images in ALL-IDB1 belonging to the same magnification are used, with 30 images used for training and 3 for testing.

\textbf{CRCHistoPhenotypes} \cite{CRC} has 100 H\&E stained images of colorectal adenocarcinoma histology with nuclei annotated on them. We use 80 images for training and 20 for testing. 

\textbf{Sigtuple WBC dataset} \cite{shonit} contains images of WBCs randomly selected from more than $1,000$ normal and abnormal peripheral blood smears prepared using May Grunwald Giemsa and Leishman stains~\cite{bancroft2008theory} imaged using a $40\times$ objective magnification in brightfield microscope. Here $5,000$ WBC patches of size $256\times256$ were used for training and $7,663$ patches of same size were used for testing.

\textbf{Training:} Adam optimizer with a learning rate of $1\times 10^{-4}$ is used. The network was trained over $500,000$ iterations updating Stages 1-4 per mini-batch. Learning rate decay by factor of 0.5 in intervals of $100,000$ iterations. Experiments were performed on a server with $2\times$ Intel Xeon 4110 CPU, $4\times32$ GB DDR4 ECC Regd. RAM, $2\times$ Nvidia Tesla V100 GPU with 16GB HBM, with software implementation on Ubuntu 16.04 LTS OS, Python 3.6, PyTorch 0.5, Nvidia CUDA 9.2 and cuDNN 7.1 for acceleration.

\begin{figure*}[t]
\begin{center}

    \begin{subfigure}[b]{0.15\textwidth}
        \centering
        \includegraphics[trim=160 355 215 40, clip, width=\textwidth]{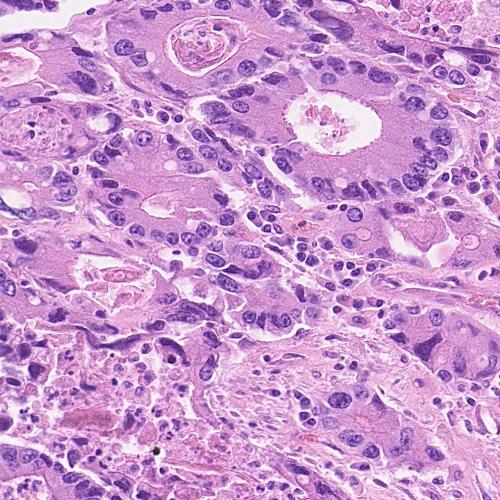}
        \caption{Real (20x objective magnification)}
        \label{fig1b}
    \end{subfigure}
    \begin{subfigure}[b]{0.15\textwidth}
        \centering
        \includegraphics[trim=160 355 215 40, clip, width=\textwidth]{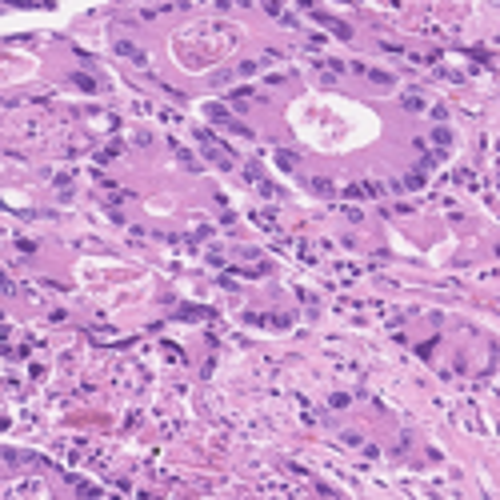}
        \caption{Bicubic \\(21.27 dB/0.49/6.81)}
        \label{fig1c}
    \end{subfigure}
    \begin{subfigure}[b]{0.15\textwidth}
        \centering
        \includegraphics[trim=160 355 215 40, clip, width=\textwidth]{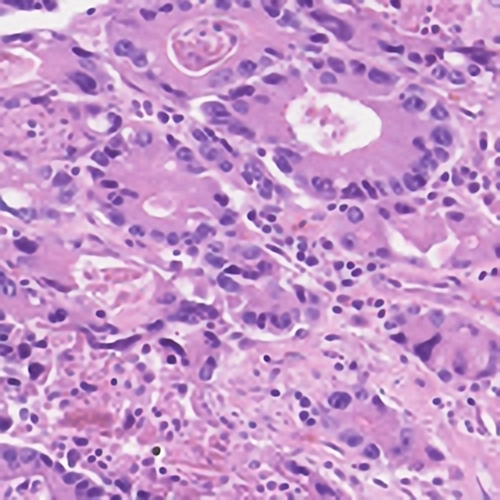}
        \caption{SRNet \\(22.17 dB/0.59/6.38)}
        \label{fig1d}
    \end{subfigure}
    \begin{subfigure}[b]{0.15\textwidth}
        \centering
        \includegraphics[trim=160 355 215 40, clip, width=\textwidth]{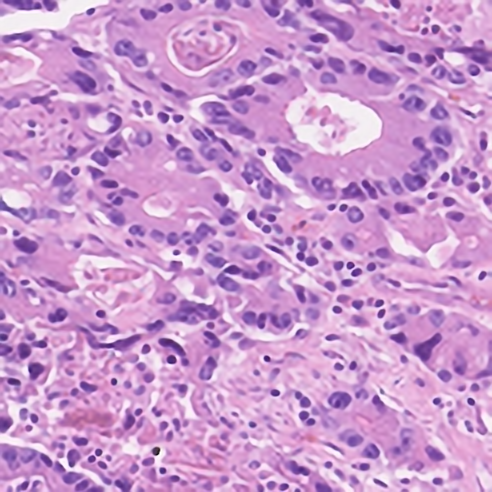}
        \caption{SRNet-w \\(22.62 dB/0.62/6.69)}
        \label{fig1e}
    \end{subfigure}
    \begin{subfigure}[b]{0.15\textwidth}
        \centering
        \includegraphics[trim=160 355 215 40, clip, width=\textwidth]{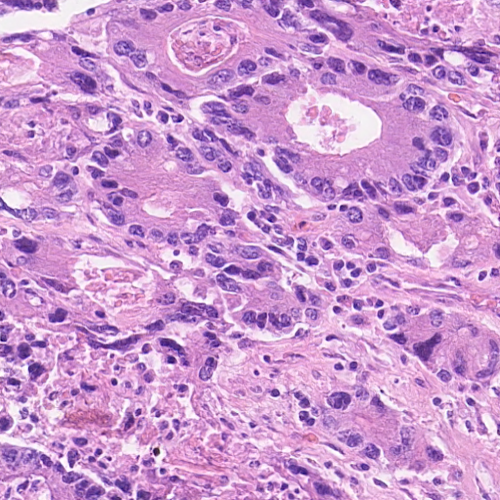}
        \caption{T1 \\(20.58 dB/0.52/3.17)}
        \label{fig1f}
    \end{subfigure}
    \begin{subfigure}[b]{0.15\textwidth}
        \centering
        \includegraphics[trim=160 355 215 40, clip, width=\textwidth]{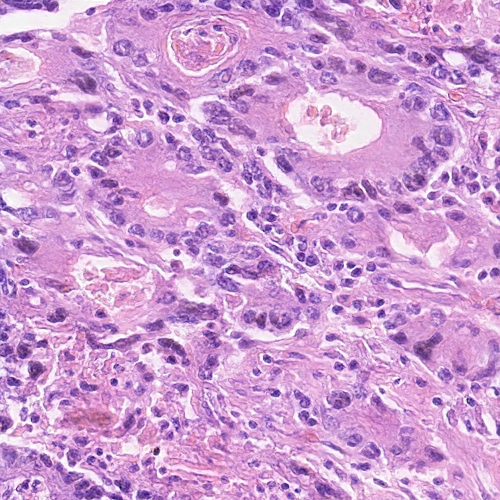}
        \caption{T1+T2 \\(19.33 dB/0.44/2.64)}
        \label{fig1g}
    \end{subfigure}
\caption{Illustration of the performance at SR factor of $16\times$ on a sample from CRCHistoPhenotypes dataset. Corresponding PSNR, SSIM and PI values mentioned in brackets.}
\label{fig:crchistophenotype}
\end{center}
\end{figure*}

\textbf{Impact of introducing rVTT} observed in the three datasets is presented in Table~\ref{table1}, where minimum loss in perceptual index\footnote{https://www.pirm2018.org/PIRM-SR.html} \cite{perception_distortion_tradeoff} with inclusion of the rVTTs is evident, along with the perception distortion tradeoff~\cite{perception_distortion_tradeoff} inline with observations in~\cite{ESRGAN}. Also observed in Fig.~\ref{fig:resultsnapshot} and Fig.~\ref{fig:crchistophenotype}.

\begin{table}[t]
\centering
\caption{Performance comparison over different datasets at SR factor of $16\times$. Higher values of PSNR and SSIM are good, lower value of Perceptual Index (PI) is good, best case marked in bold. SRNet-w has edge weighting in $J_{Recon}()$.}
\label{table1}
\footnotesize
\begin{tabular}{lrrrrrr}
\textbf{ALL} & Nearest & Bicubic & SRNet & SRNet-w & T1 & T1+T2 \\
\hline
PSNR & 32.83 & 37.65 & 38.64 & \textbf{43.03}  & 32.77 & 37.98 \\
SSIM & 0.88 & 0.94 & 0.95 & \textbf{0.97} & 0.92 & 0.96 \\PI  & 7.27 & 8.15 & 7.2 & 7.32 & 5.95 & \textbf{5.31}  \\
 [0.3cm]

\textbf{CRCH} & & & & & &  \\
\hline
PSNR & 22.41 & 25.26 & 25.91 & \textbf{26.31}  & 24.74 & 23.34 \\
SSIM & 0.57 & 0.63 & 0.69 & \textbf{0.71} & 0.64 & 0.59 \\PI  & 13.07 & 7.38 & 6.53 & 7.19 & 3.71 & \textbf{3.27}\\
 [0.3cm]
\textbf{ST} & & & & & &   \\
\hline
PSNR & 24.83 & 30.1 & \textbf{36.5} & 36.33  & 34.93 & 34.61  \\
SSIM & 0.78 & 0.88 & \textbf{0.96} & 0.95 & 0.94 & 0.94\\PI  & 7.34 & 8.02 & 7.19 & 7.28 & 7.07 & \textbf{6.52} \\
 [0.3cm]
\end{tabular}
\end{table}

\begin{figure*}[t]
 \centering
    \begin{subfigure}[b]{0.18\textwidth}
        \centering
         \includegraphics[trim=84 80 80 80, clip, width=\textwidth]{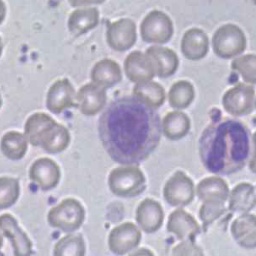}
        \caption{Original Image (40x \\objective magnification)}
        \label{fig:Original Image}
    \end{subfigure}
    \begin{subfigure}[b]{0.18\textwidth}
        \centering
        \includegraphics[trim=84 80 80 80, clip, width=\textwidth]{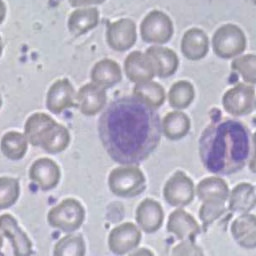}
        \caption{4x \\(41.65 dB/0.98/6.35)}
        \label{fig:4x SR}
    \end{subfigure}
    \begin{subfigure}[b]{0.18\textwidth}
        \centering
        \includegraphics[trim=84 80 80 80, clip, width=\textwidth]{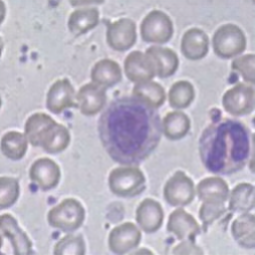}
        \caption{9x \\(33.42 dB/0.93/6.05)}
        \label{fig:9x SR}
    \end{subfigure}
    \begin{subfigure}[b]{0.18\textwidth}
        \centering
        \includegraphics[trim=84 80 80 80, clip, width=\textwidth]{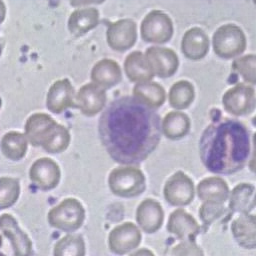}
        \caption{16x \\(34.41 dB/0.94/5.95)}
        \label{fig:16x SR}
    \end{subfigure}
    \begin{subfigure}[b]{0.18\textwidth}
        \centering
        \includegraphics[trim=84 80 80 80, clip, width=\textwidth]{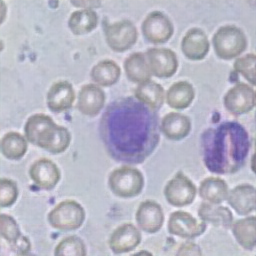}
        \caption{64x \\(27.86 dB/0.75/6.19)}
        \label{fig:64x SR}
    \end{subfigure}

    \caption{Effect of scale of SR on image appearance for a sample in Sigtuple WBC Dataset. Corresponding PSNR, SSIM and PI.}
    \label{scale_fig}
\end{figure*}

\textbf{Role of rVTTs across scale of super resolution} is visible as the scale increases from $4\times$ to $64\times$, in Fig.~\ref{scale_fig}.

\textbf{Equivocal diagnosis with use of SR} is demonstrated using a commercially available artificial intelligence (AI) digital pathology system\footnote{https://sigtuple.com/\#s-solutions}~\cite{shonit} for inferring with the Sigtuple WBC Dataset with its results presented in Table~\ref{table2}. This justifies role of $T1(\cdot)$ and $T2(\cdot)$ in restoring texture of diagnostic importance beyond what can be achieved using simple interpolation on learning a SRNet without rVTT adversarial framework. This proves diagnostic equivocality of SR with rVTT to serve matching purpose as Real HR images while reducing the image acquisition time, speeding up the diagnosis delivery time.

\begin{table}[t]
\centering
\caption{Overlap of AI based diagnosis~\cite{shonit}  (in \%) using interpolated and SR images against using Real HR ground truth.}
\label{table2}
\footnotesize
\begin{tabular}{lllllll}
Scale & Nearest & Bicubic &SRNet & SRNet-w & T1 & T1 + T2\\
\hline
4x & 97.56 & 98.15  & 97.79 & 97.81  & 99.36 &\textbf{99.50}\\
9x & 85.55 & 96.11  & 97.79 & 97.44 & 98.29 &\textbf{98.58}\\
16x  & 81.43 & 93.01  & 97.26 & 96.53  & \textbf{98.15} &97.81\\
 [0.3cm]

\end{tabular}
\end{table}

\section{Conclusion}
Here we have proposed using two rVTTs for enhancing the performance of a SRNet for microscopy with a specific focus on being able to restore the texture within diagnostically relevant nucleus and around cytoplasm in cells. We demonstrate the marked rise in performance of the SRNet with this arrangement in line with philosophy of~\cite{ESRGAN} also proving its equivocal response similar to a Real HR image when used for inferencing with an AI based digital pathology system. The quality of the super-resolved images evaluated using recent advancement in understanding of distortion and perception~\cite{perception_distortion_tradeoff} based measures~\cite{perceptual} also advocates in support of our claim to be able to super resolve with pathorealism retained. 

\small
\bibliographystyle{IEEEbib}
\bibliography{refs}

\end{document}